\documentclass{article}

\usepackage[dblblindworkshop, final]{neurips_2025}
\workshoptitle{Bridging Language, Agent, and World Models for Reasoning and Planning}

\usepackage[utf8]{inputenc}
\usepackage[T1]{fontenc}
\usepackage{hyperref}
\usepackage{placeins}
\usepackage{url}
\usepackage{booktabs}
\usepackage{amsfonts}
\usepackage{amsmath}
\usepackage{amssymb}
\usepackage{nicefrac}
\usepackage{microtype}
\usepackage{xcolor}
\usepackage{graphicx}
\usepackage{subfigure}
\usepackage{mathtools}
\usepackage[capitalize,noabbrev]{cleveref}
\usepackage{amsthm}
\usepackage{tabularx}

\usepackage{caption}             
\usepackage{subcaption}          
\usepackage{float}               

\theoremstyle{plain}

\theoremstyle{definition}

\theoremstyle{remark}

\title{Causal Masking on Spatial Data: An Information-Theoretic Case for Learning Spatial Datasets with Unimodal Language Models}

\author{%
  Jared Junkin\thanks{Equal contribution.} \\
  Department of Electrical and Computer Engineering\\
  Johns Hopkins University\\
  Baltimore, Maryland, United States \\
  \texttt{jjunkin2@jh.edu}
  \And
  Samuel Nathanson \\
  Whiting School of Engineering \\
  Johns Hopkins University \\
  Baltimore, Maryland, United States \\
  \texttt{samuel.nathanson@jhu.edu}
}

\begin{document}

\maketitle

\begin{abstract}
Language models are traditionally designed around causal masking. In domains with spatial or relational structure, causal masking is often viewed as inappropriate, and sequential linearizations are instead used. Yet the question of whether it is viable to accept the information loss introduced by causal masking on nonsequential data has received little direct study, in part because few domains offer both spatial and sequential representations of the same dataset. In this work, we investigate this issue in the domain of chess, which naturally supports both representations. We train language models with bidirectional and causal self-attention mechanisms on both spatial (board-based) and sequential (move-based) data. Our results show that models trained on spatial board states - \textit{even with causal masking} - consistently achieve stronger playing strength than models trained on sequential data. While our experiments are conducted on chess, our results are methodological and may have broader implications: applying causal masking to spatial data is a viable procedure for training unimodal LLMs on spatial data, and in some domains is even preferable to sequentialization. 
\end{abstract}

\section{Introduction}
\label{Introduction}
Causal masking enforces left-to-right next-token prediction and reflects the inherently sequential structure of natural language. In domains with spatial or relational data it is desirable to utilize other forms of attention which do not apply causal masking. 

Natively multimodal models employ a variety of techniques to preserve causal masking across sequential data while allowing for the application of methods such as bidirectional attention, cross-attention bridges, and fused attention to non-sequential modalities. However, natively multimodal models are substantially more complicated and expensive to train due to the heavy compute and memory overhead of cross-modal attention, the difficulty of curating and balancing large paired datasets across modalities, and the burden of building modality-specific tokenization and embedding methods. This means it is  desirable to explore means by which training can be made simpler or avoided altogether. This raises the question of whether it is possible to accept the information loss resulting from applying causal masking to spatial data, training a unimodal autoregressive LLM as if it were ingesting sequential data. 

This question has largely gone unaddressed, due to both the existence of natively multimodal models and the fact that there are few domains where equivalent spatial and sequential representations of data exist -- a necessary prerequisite for foundational research to occur, due to the need for benchmarking and quantitative analysis of results. Chess is an ideal domain in which to examine this question. We believe the recent emphasis on smaller, cheaper language models which sacrifice exhaustive knowledge of the world in favor of greater reasoning abilities further motivates this as a timely area of research. 

\subsection{Learning a Transformer-Based Chess Agent}

Chess offers two textual encodings: PGN (sequential moves) and FEN (spatial board state with side-to-move/rights/counters; see Appendix~\ref{appendix:notation}). We use both; Figure ~\ref{fig:pgn_notation} illustrates them for a R\'uy Lopez position.

\begin{figure}[t]
  \centering
  \includegraphics[width=\textwidth]{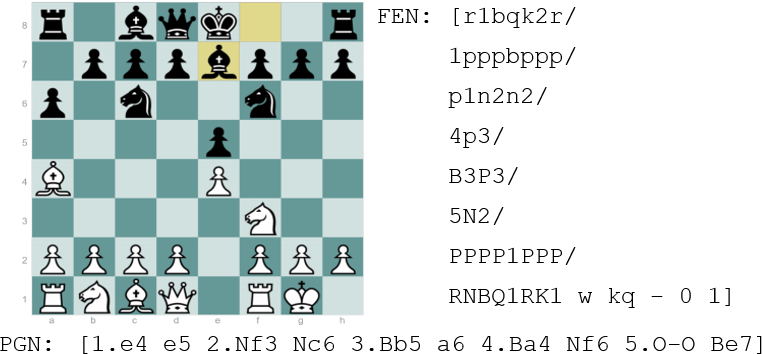}
  \caption{PGN vs.\ FEN encodings for a R\'uy Lopez position.}
  \label{fig:ruylopez_pgn_fen}
\end{figure}

\subsubsection{\textbf{Learning from PGN Data}}
PGN data is a sequential (causal) list of moves made over the course of a chess game starting with move \(m_0\) and ending with move \(m_n\) for a game lasting \(n\) moves. Because of this, PGN can be interpreted  the same way as natural language, and a causal autoregressive LLM \(\Pi_\theta\) can be learned on PGN that maximizes the log probability of correctly predicting the ground truth next move \(m_t\) at each timestep \(t \in T\) given the sequence of preceding moves made (\(m_0, m_1, \ldots, m_{t-1}\)) for each sample \(s\) in some PGN Dataset \(\mathcal{D}\):
\[
\hat{m}_t = \arg\max_{m \in \mathcal{M}} \left[ \log P(\hat{m}_t ==m_t\mid m_0,m_1, \ldots, m_{t-1})\right]
\]
A categorical cross-entropy loss is imposed during training, defined as:
\[
\mathcal{L} = -\sum_{i=1}^{T} \sum_{j=1}^{M} \mathbb{I}(y_{ij}==m_i)\log(y_{ij})
\]
Where \( T \) gives the number of moves in a given game, \( \mathbb{I}\) gives the indicator function, and \(M\) represents the size of the vocabulary space under the the given tokenization strategy. As  \(\mathcal{L} \rightarrow 0\), the agent's behavior approximates that of the player(s) who generated the dataset. Formally, if we assume our dataset  \(\mathcal{D}\) was generated by some oracle \(\Upsilon\) which decides which move \(m_t\) to play at timestep \(t \in T\) given a PGN string \(s\) representing moves \(m_0, m_1, \ldots, m_{t-1}\), then for a well-trained model \(\Pi_\theta\):
\[\Pi_\theta(s) \approx \Upsilon(s)\]
It's worth noting that this means the model's skill is upper-bounded by the skill of the oracle \(\Upsilon\). 

\subsubsection{\textbf{Learning from FEN Data}}
FEN is non-sequential; we train a classifier over legal moves with cross-entropy on the best move \(m^\ast\) (full objective in Appendix~\ref{appendix:objective}).


\section{Related Work}
\label{sec:relatedwork}
Transformer-based chess agents have followed two main paths. The first fine-tunes pretrained LLMs on chess formats: GPT-2 on PGN \citet{ciolino,gpt2,pythia2023}, anecdotal GPT-3.5 evaluations \citet{acher}, and LLaMA-style models fine-tuned on FEN to reach professional strength \citet{zhang2025chessllm}. These show feasibility but remain relatively rare compared to bespoke models.

A second line trains small decoder-only transformers from scratch. Karvonen \citet{karvonen} (PGN) and OthelloGPT \citet{hazineh2023othello} indicate that autoregressive transformers learn latent spatial encodings from sequential moves. Ruoss et al.\ introduce ChessBench and strong FEN-based models with engine annotations \citet{chessbench,ruoss2024grandmaster}. Larger-scale efforts combine strong policies with search or alternative objectives across board games \citet{schultz2025boardgames,hamara2025planning,ye2025diffusearch}. In parallel, interpretability work shows structured world-model representations and belief-state geometry in transformers trained on symbolic/spatial tasks \citet{tosh,ivanitskiy2023maze,spies2024maze,shai2024belief}.

Masking strategies in spatial or multimodal settings often relax pure causal masking (e.g., block-causal for images; relaxed masks in V+L) \citet{amrani2025xtra,pei2025rethinking}. By contrast, our contribution is methodological: we directly compare \emph{causal-masked} training on spatial FEN to the conventional sequentialization path (PGN), and we show that accepting the information loss from causal masking on spatial inputs is preferable to linearizing spatial data for causal models. Additional survey details and a comparison table are provided in Appendix~\ref{app:relatedwork}.

\section{Methods}
\subsection{Formalization of Problem}
To begin with, we offer a brief formal justification for our hypothesis that training on spatial chess data is more desirable than training on sequential chess data. A model \(\Pi^P_\theta\) trained on PGN will learn to approximate the ground truth optimal policy:
\[\Pi^P_\theta(m_0, m_1 \ldots m_{t-1}) \approx \Pi^\ast(m_0, m_1 \ldots m_{t-1}) \]

Meanwhile, a model \(\Pi^F_\theta\) trained on FEN will learn to approximate a first-order Markovian policy \(\Pi_{\text{Markov}}\) in which the best next move \(m^\ast\) is determined exclusively by the current board state \(B_t\):

\[\Pi^F_\theta \approx \Pi_{\text{Markov}}(B_t) = \arg \max_{m \in M} \left[Q^\ast(m | B_t)\right]\]

Because chess is played with perfect information and zero-stochasticity, it is first-order Markovian with the exception of the draw by threefold repetition rule, which states that if the same board position is repeated three times, the game ends in a draw. From a mathematical perspective, if this rule did not exist, chess would obey the Markov Property, which states that the future is conditionally independent of the past given the present:
\[P(s_{t+1} | s_t, a_t, s_{t-1}, a_{t-1}, \ldots s_0, a_0) = P(s_{t+1} | s_t, a_t)\]

However, because of the draw by threefold repetition rule, the true optimal policy of chess is dependent on the entire sequence of board states \((B_0, B_1, \ldots B_{t-1})\) so that \(\Pi^\ast\) can avoid draws by threefold repetition in games where a winning sequence of moves exists: 
\[\Pi^\ast(B_0 \ldots B_t) = \arg \max_{m \in M} \left[Q^\ast(m_i | (B_0 \ldots B_t)\right]\]
It's also possible to express an optimal policy for determining the optimal move \(m^\ast\) at timestep \(t\) in terms of the sequence of moves previously made \((m_0, m_1 \ldots m_{t-1})\), because the sequence of board states can be extracted without loss of information from the sequence of moves: 
\[\Pi^\ast(m_0, m_1 \ldots m_{t-1}) = \arg \max_{m \in M} \left[Q^\ast(m_i |(m_0, m_1 \ldots m_{t-1})\right]\]
Because FEN data doesn't contain any information on past board states, any model \(\Pi_\theta\) learned with FEN data will learn to approximate \(\Pi_{\text{Markov}}\):
\[\Pi_\theta(B_t) \approx \Pi_{\text{Markov}} \neq \Pi^\ast(m_0, m_1 \ldots m_{t-1})\]

For a discussion of how this impacts model behavior and our solution, please see the appendix. Despite the fact that a model \(\Pi^F_\theta\) trained on FEN will learn to approximate a suboptimal policy, we found that models trained on FEN data exhibited superior performance in practice. 

Let \(\mathcal{P}\) denote the space of all possible PGN strings (i.e., all sequences of valid moves from the start of a standard chess game) and \(\mathcal{F}\) denote the set of all possible FEN strings. We treat \(\mathcal{P}\) and \(\mathcal{F}\)  as  countably infinite sample spaces from which we draw samples \(p \in \mathcal{P}\) and \(f \in \mathcal{F}\) during training. Our models \(\Pi_\theta^P\) and \(\Pi_\theta^F\), models parameterized by \(\theta\) and trained on PGN and FEN data respectively, learn mappings from \(\mathcal{P}\) and \(\mathcal{F}\) to the finite set \(\mathcal{S}\), which represents all possible SAN (Standard Algebraic Notation) moves:
\[\Pi_\theta^P: \mathcal{P} \rightarrow \mathcal{S}\]
\[\Pi_\theta^F: \mathcal{F} \rightarrow \mathcal{S}\]
For a given PGN string \(p \in \mathcal{P}\) there exists a surjective (many-to-one) mapping function \(\mathcal{G}\) which will convert \(p \in \mathcal{P} \) into a unique FEN string \(f \in \mathcal{F}\):
\[\mathcal{G}: \mathcal{P} \rightarrow \mathcal{F}\]
\[\mathcal{G}(p_i) = f_i\]
The relationship between pieces on a chessboard (which pieces threaten which, which pieces guard which) is determined by their spatial orientation to each other. For example, bishops threaten pieces along their active diagonal; rooks threaten pieces along the file or row. Because of this, a policy for choosing the best move \(\Pi^\ast\) must condition on \emph{spatial} information. While the \emph{data} in PGN is structured sequentially, the ground truth information the model must reason against when evaluating move choices is \emph{structured spatially}. Therefore a model \(\Pi_\theta^P\) must back out some latent spatial representation of the board state \(B\) from the PGN sequence \(p \in \mathcal{P}\) in order to generate the best move \(s \in \mathcal{S}\); that is, a model \(\Pi_\theta^P\) trained on PGN data actually learns a composition of functions: 
\[\Pi_\theta^P: \mathcal{G} \circ (\mathcal{F} \rightarrow \mathcal{S}) \]
Where \(\mathcal{G}\) is the surjective mapping of PGN data into the FEN space (or, more specifically, into some latent dimension approximating a spatial reconstruction of board state \(B\)), while a model \(\Pi_\theta^F\) trained from FEN data does not have to perform this composition, as the information is already structured spatially:
\[\Pi_\theta^F: \mathcal{F} \rightarrow \mathcal{S}\]
We hypothesize that the composition \(\mathcal{G} \circ (\mathcal{F} \rightarrow \mathcal{S})\) entails greater representational complexity than the direct mapping from FEN to SAN moves \(\mathcal{F} \rightarrow \mathcal{S}\). Intuitively, PGN-based models must internally reconstruct spatial board states before selecting moves, whereas FEN-based models can condition directly on spatial structure. We present this as an information-theoretic intuition rather than a formal proof.

\subsection{Models \& Datasets}
We conduct our experiments on Meta AI's 1.3B Parameter Llama3.1 Model \citet{llama3tokenizerpaper} and two identically-sized character-level language models. One character-level language model was trained on PGN data with causal masking, and the other was trained on FEN data without causal masking; meanwhile the Llama model was trained using FEN data \emph{while using causal masking}. For our FEN dataset, we utilized the Chessbench dataset provided by \citet{chessbench}, which consists of 15 billion board positions annotated with the top move according to Stockfish \citet{stockfish}, which is the world's strongest chess engine as of January 2025. Our PGN dataset consisted of approximately 1 Billion PGN strings representing games played between a maximum-strength Stockfish agent as white and a variety of attenuated chess engines with ELO ratings between 1200 (advanced beginner) and 3100 (superhuman skill) playing as black. The dataset was assembled this way to ensure that neither moves played by beginners nor experts were out-of-distribution. Our PGN model only played as white in the simulated tournaments we utilized for evaluation, while both the models trained on FEN played both white and black. 

\subsection{Prompting \& Tokenization}

For FEN, we enforce character-level tokenization to avoid ambiguous merges and align with the pretrained vocabulary; prompts embed the FEN, the legal SAN moves, and the engine best move, which stabilizes training. Details on merges/run-length handling, templates, and padding appear in Appendix~\ref{appendix:tokenization}.

\textbf{Full tokenization templates, padding details, and examples are provided in Appendix~\ref{appendix:tokenization}.}

\subsection{Objective Function}
Let \(X\) be the tokenized prompt described above. Let the tokens in this prompt representing the best move be denoted by by \(m^{\ast} = (m^{\ast}_{1}, \ldots, m^{\ast}_{k})\), and let the tokens we predict for the best move be denoted as \(\hat{m} = (\hat{m}_1, \hat{m}_2 \ldots \hat{m}_k)\). We create a binary loss mask \(\mathbf{w} = (w_{1}, \ldots, w_{T})\) of length \(T = |X|\), where
\[
   w_{t} \;=\; 
   \begin{cases}
     1, & \text{if token } t \text{ is part of } m^{\ast},\\
     0, & \text{otherwise}.
   \end{cases}
\]
Let the model’s predicted distribution at step \(t\) be 
\[
   p_{\theta}\bigl(y_{t} \,\mid\, X\bigr),
\]
where \(\theta\) are the model parameters. Our objective function is a masked cross-entropy loss wherein we only sum over tokens belonging to \(m^{\ast}\). All other predictions are masked out (where \(w_{t} = 0\)), contributing no penalty to \(\mathcal{L}\):
\[
   \mathcal{L}_{\text{masked}}(\theta) =- \sum_{t=1}^{T}  w_{t}      \log p_{\theta}\Bigl(m^{\ast}_{t} \Bigm|X_{0:t-1}\Bigr),
\]
We use a typical lower-triangular causal attention mask, so that at step \(t\), the model can only attend to tokens up to \(t-1\). Padding tokens are also masked.  It's worth emphasizing that because the ground truth target tokens \((m^\ast_{<t} = m^\ast_0 \ldots m^\ast_{t-1})\) are also a part of our tokenized prompt \(X_{0:t-1}\), our objective function is fully \emph{teacher forced} \citet{teacherforcing}. The final gradient update to \(\theta\) is thus driven solely by errors in predicting the best-move tokens, and each target token \(m^\ast_i\) conditions on the full prompt (including the target tokens preceding it). 

\noindent \textbf{Exposure bias (brief).} We observed only modest differences between teacher-forced and autoregressive decoding; full metrics and figures are provided in Appendix~\ref{appendix:exposurebias}.


\section{Experiments}

\subsection{Training Details}
All models trained for 200{,}000 steps on 2$\times$A100 (80GB) with cosine LR decay, warmup, gradient clipping, and mixed precision. We followed published hyperparameters for Llama and Karvonen-style settings for the character baselines. Results are from a single training run per game due to compute constraints (each run $\approx$3 weeks on our hardware). Full configurations and loss curves are reported in Appendix~\ref{appendix:training} and Appendix~\ref{appendix:losscurves}.

\subsection{Empirical Validation that Training on Spatial Data Produces Superior Skill}
Our first experiment is an empirical justification of the hypothesis formulated in the previous section. As mentioned in the introduction, we trained two identical NanoGPT models on equivalently sized datasets of PGN and FEN data. The only difference between these two models is that the model trained on PGN utilized causal self-attention with lower-triangular masking, while the model trained on FEN utilized bidirectional attention with no masking, allowing it to attend to all tokens simultaneously because FEN data is spatial. We trained both models for 200,000 training steps with equal batch sizes and hyperparameter values. We then evaluated the mean cross-entropy per sample for FEN and PGN test data:\\

\[
\text{CE}_{\text{PGN}} = -\frac{1}{N}\sum_i\sum_{t \in |p_i|}\log P_{M^P_\theta}(\hat{m}_{it} | p_i) 
\]

\[
\text{CE}_{\text{FEN}} = -\frac{1}{N}\sum_i\log P_{M^F_\theta}(\hat{m}_i | f_i) 
\]

Where \(\hat{m}_i\) gives the model's predicted move for the \(i^{th}\) sample. The equation for PGN contains a second sum because a single move will contain \(t\) tokens; in FEN data, the tokenization scheme represents each move as a single token. 

\begin{figure}[t]
    \centering
    \includegraphics[width=\textwidth]{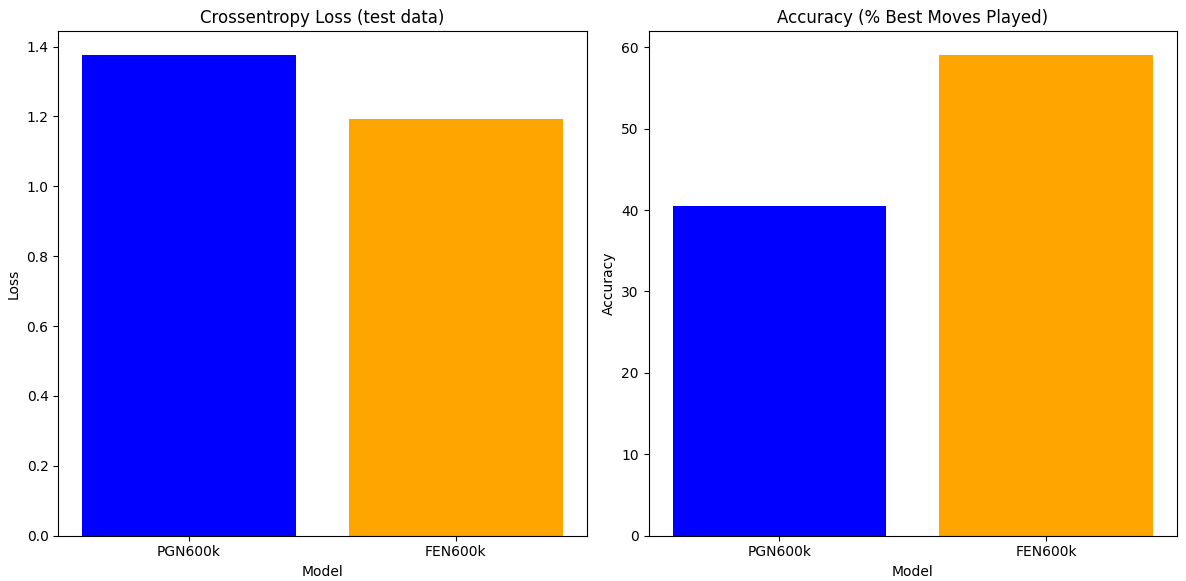}
    \caption{The total cross entropy loss per prediction (left) and accuracy in \% of top engine moves played (right) for models trained on FEN (blue) and PGN (yellow)}
    \label{fig:pgn_notation}
\end{figure}

We observed that the model trained on PGN did indeed have substantially higher loss values and lower best-move classification when compared to the model trained on FEN. 

\subsection{Causal Masking on Spatial Data}
Next we applied the same training regimen to our causal Llama model, training it to play chess via a large-scale supervised fine tuning run on our FEN dataset. The model displays grandmaster-level performance with an estimated ELO rating of 2630, approximately 500 points higher than that of the causal model trained on PGN and only narrowly worse than the bidirectional model trained on FEN. These results demonstrate that when training causal language models to play chess, it is better to accept the information loss from applying causal masking to sequential FEN data than to linearize the data into a sequential representation beforehand. 

On a held-out set of \(\sim 12{,}800\) positions (zero-shot baseline compared to performance after SFT):

\begin{itemize}
    \item \textbf{Syntactically valid move rate:} \(99.94\%\) after SFT.
    \item \textbf{Legal move rate:} \(99.91\%\) after SFT.
    \item \textbf{Best move (Stockfish):} \(\sim 0.6\% \rightarrow \approx 58\%\) (\(\sim\!100\times\) increase).
\end{itemize}


\begin{figure}[t]
    \centering
    \includegraphics[width=\textwidth]{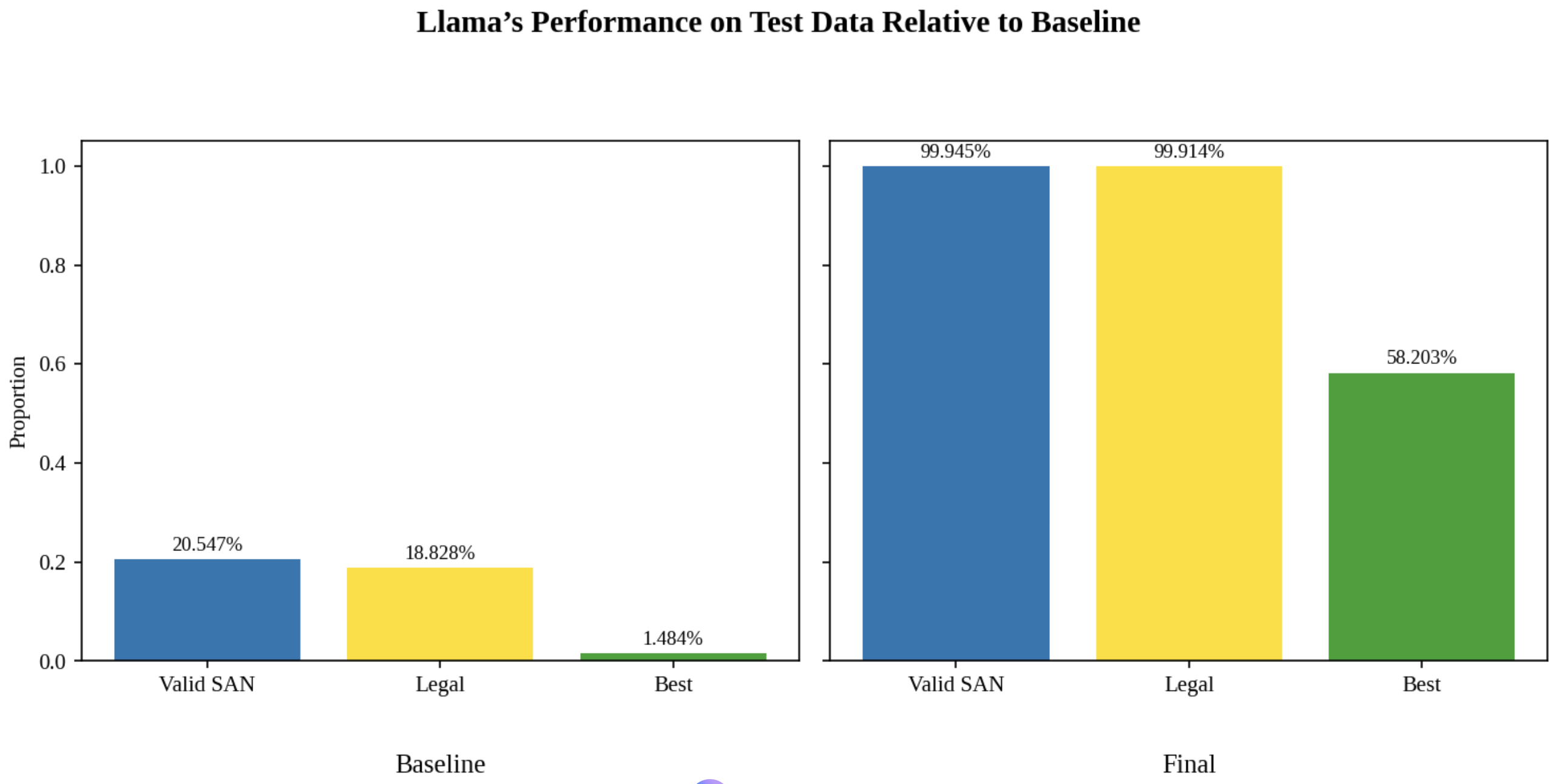}
\caption{The \% of prompts resulting in Syntactically Valid, Legal, and Best (according to chess engine) moves. \emph{Note:} All numbers are from a single run per game. See Appendix \ref{appendix:training} for hardware/runtime.
}
\end{figure}

We then compared this to both of our other models' inference-time performance on the same three metrics (there was no sense baselining either of the other two models, as we trained them from scratch rather than conducting SFT). All three models are capable of outputting valid and legal moves with nearly 100\% accuracy; however, there is a clear gradient in their abilities to output the best Stockfish move. The model trained on PGN outputs the best move only 40\% of the time, while our Llama model outputs the best move 58\% of the time and the bidirectional model trained on FEN outputs the best move over 60\% of the time.

\begin{figure}[t]
    \centering
    \includegraphics[width=\textwidth]{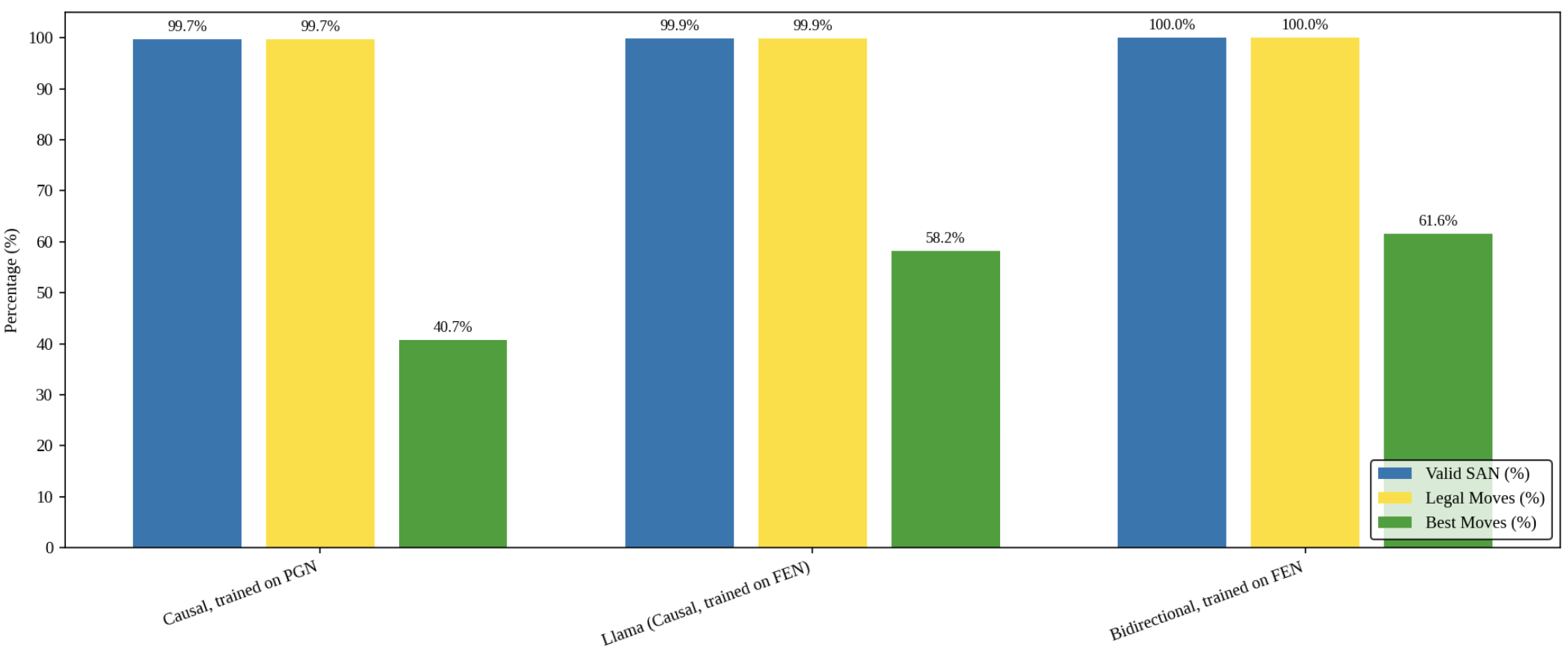}
\caption{The Percentage of Prompts resulting in Syntactically Valid, Legal, and Best (according to chess engine) Moves.}
\label{fig:all_performance}
\end{figure}



\subsection{Calculating Elo Rating}
We estimated Elo by playing each model against calibrated Stockfish agents (Levels 0–10), using 1000 games per level (balanced White/Black) and standard Elo updates. We report headline Elo in the main text; full protocol, equations, and the Stockfish calibration table appear in Appendix~\ref{appendix:elo} and Appendix~\ref{appendix:stockfish_ratings}.

\section{Discussion}
This paper presents a case study of a causal-masked LLM that demonstrates grandmaster-range performance under Elo evaluation against Stockfish agents and demonstrates that, when training a casual language model to play chess, it is better to accept the information loss that results from applying causal masking to spatial data than to linearize board state into a sequential representation ahead of causal masking. We have presented a compelling theoretical case that this is due to the higher functional complexity resulting from having to back out latent representations of relevant spatial features in the board state, and backed it up with an empirical analysis. With a high-quality dataset, sophisticated prompting and tokenization strategies, and adherence to published hyperparameter values and best practices training principles, it is possible to learn highly sophisticated domain-specific behavior from non-causal data with causal LLMs -- even with small LLMs and modest compute resources. We believe that abstract strategy games such as chess remain incredibly fertile ground for AI research, and hope this result stimulates future research across other domains in machine learning. To that end, we conclude our paper with several avenues we think warrant further exploration, as well as key lessons learned which we hope will aid future researchers: 

\subsection{\textbf{Scaling Laws, Pretraining Quality, \& Overparameterization}}
We also experimented with training a 140M parameter Pythia model to play chess. We observed Llama consistently outperformed Pythia even when the two models were trained on equivalent batch sizes for identical numbers of training steps. 

However, both models are  \emph{overparameterized} for the task of learning chess, because Ruoss \citet{chessbench,ruoss2024grandmaster} and Karvonen \citet{karvonen} both published results with 50M parameter models that reached high skill levels in chess. This means that either \emph{A)} the quality and duration of pretraining is a significant determinant in model performance, or \emph{B)} the number of non-embedding parameters in a LLM is a significant determinant in performance \emph{even when} both models are overparameterized relative to the problem. Exploring which of these, or both, is the case, would be an interesting avenue to for future resesarch.

\subsection{\textbf{Chess \& LLM Interpretability}}
Abstract strategy games like chess have a far smaller vocabulary size than natural language, intuitive visual interpretations, and quantitative means of assessing response quality and correctness. Because of this, it's an extremely interesting sandbox for evaluating and developing new methods for AI interpretability.

\subsection{\textbf{Importance of Tokenization Process}}
In our experiments, we found that tokenization strategy had a significant effect on training convergence. Both the Pythia and Llama models failed to converge under their default tokenizers, which merge character sequences such as ``pk'' (pawn–king) into a single token. To address this, we modified the tokenizer to operate strictly at the character level and flattened run-length encodings in FEN strings. These changes ensured that each chess piece and empty square was represented consistently, and that counters were aligned across all samples. After this modification, models trained stably and achieved strong performance. We interpret this result as evidence that a well-matched tokenization schema is an important factor when adapting pretrained models to highly structured symbolic domains such as chess. While we cannot rule out that alternative strategies might also succeed, our findings highlight that careful preprocessing is often necessary to align domain-specific inputs with the statistical assumptions of pretrained tokenizers.

\section{Limitations}
Our work has several limitations. First, Elo ratings were computed against attenuated Stockfish agents; while these provide a calibrated baseline, they are not directly equivalent to FIDE Elo ratings for human players. Second, models trained on FEN data are blind to the threefold repetition rule, meaning they cannot avoid certain draws without additional mechanisms. Third, our results depend on a single training run per condition with modest compute, limiting variance estimates. Finally, while we hypothesize broader implications of applying causal masking to spatial domains, our empirical results are confined to chess; generalization to other domains remains future work.

\section{\textbf{Conclusion}}
Chess has been an active area of AI research for so long that the game is commonly referred to as "the Drosophila of artificial intelligence." In this work, we introduced, to our knowledge, the first open-source fine-tuned LLM to achieve performance in the $\sim$2600 Elo range when calibrated against Stockfish agents. More importantly, we provided a methodological case study showing that causal masking on spatial data, while lossy, can outperform sequentialization, and that careful tokenization and prompting strategies allow pretrained LLMs to exploit structured symbolic domains effectively.  

Our first key insight is that, because the relationships between pieces on a chess board are spatial, any model
\[
\Pi_\theta^P: \mathcal{P} \rightarrow \mathcal{S}
\]
trained to output chess moves from sequential PGN inputs must still recover spatial board-state information. While this information exists in compressed form within PGN and can be extracted losslessly, the model must devote additional capacity to forming a latent spatial representation
\[
\mathcal{G} \circ (\mathcal{F} \rightarrow \mathcal{S}),
\]
which approximates this reconstruction. Intuitively, this mapping entails greater representational complexity than training a model
\[
\Pi_\theta^F: \mathcal{F} \rightarrow \mathcal{S}
\]
directly on spatial FEN data. We hypothesize that this explains why, in practice, FEN-trained models tend to achieve stronger performance, even though they operate under causal masking and lose some bidirectional information.  

Our second key insight was the importance of aligning tokenization and prompting with the symbolic structure of the domain. By enforcing character-level tokenization for FEN and embedding domain-specific information directly into prompts, we ensured stable training and enabled LLMs to represent board states consistently. This highlights how tokenizer alignment is not a technical afterthought, but a central design choice when adapting pretrained models to structured domains.  

Our findings highlight chess as a fertile and interpretable sandbox for studying representation learning, masking strategies, and world-model formation in LLMs. While our results are confined to chess, the methodological lessons---particularly around spatial causal masking and tokenizer alignment---may extend to other spatially structured domains. For the LAW community, we view this as evidence that causal masking can remain a viable design choice for structured reasoning tasks, and that abstract games provide a uniquely transparent setting in which to probe questions about world models and representation.  

We hope that this study not only contributes a strong open-source baseline for chess-playing LLMs, but also motivates further exploration of causal masking and representation learning in broader spatial reasoning contexts.  

\bibliographystyle{plainnat}
\bibliography{neurips_references}

\appendix

\newpage

\section{Expanded Related Work}
\label{app:relatedwork}

\subsection{Narrative overview}
We group prior work into five themes. (i) \textbf{Fine-tuning pretrained LLMs on chess} shows feasibility but is relatively sparse: GPT-2 on PGN \citet{ciolino,gpt2,pythia2023}, anecdotal GPT-3.5 strength \citet{acher}, and LLaMA-style models fine-tuned on FEN to professional levels \citet{zhang2025chessllm}. (ii) \textbf{Small models from scratch} demonstrate that even nano-scale GPTs trained on PGN or FEN can learn latent spatial structure and achieve strong play \citet{karvonen,hazineh2023othello,chessbench}. (iii) \textbf{Scaling to grandmaster-range} uses larger models, engine annotations, and/or search to reach superhuman strength \citet{ruoss2024grandmaster,schultz2025boardgames,hamara2025planning,ye2025diffusearch}. (iv) \textbf{Interpretability/world models} finds that transformers trained on symbolic/spatial domains encode legal-state tracking, causal world models, and belief-state geometry \citet{tosh,ivanitskiy2023maze,spies2024maze,shai2024belief}. (v) \textbf{Masking strategies} adapt or relax causality for spatial inputs \citet{amrani2025xtra,pei2025rethinking}. 

Our work differs by \emph{explicitly} comparing (a) PGN sequentialization under causal masking and (b) \emph{spatial} FEN under causal masking, showing the latter outperforms despite information loss from masking.

\begin{table}[H]
\centering
\small
\renewcommand{\arraystretch}{1.25}
\setlength{\tabcolsep}{4pt}
\caption{Representative prior work across five themes. ``Masking'' indicates the dominant attention style in the cited setup.}
\label{tab:related_work_compact}
\begin{tabularx}{\textwidth}{@{}p{2.8cm} X p{2.2cm} p{2.1cm} X@{}}
\toprule
\textbf{Theme} & \textbf{Representative works} & \textbf{Representation} & \textbf{Masking} & \textbf{Key takeaway} \\
\midrule
Fine-tuning LLMs &
GPT-2 on PGN {ciolino,gpt2,pythia2023}; GPT-3.5 notes \citet{acher}; LLaMA on FEN \citet{zhang2025chessllm} &
PGN / FEN &
Causal (PGN), mixed (FEN) &
Feasible to adapt pretrained LLMs; FEN fine-tuning can reach pro-level strength. \\

Small models from scratch &
Karvonen (PGN) \citet{karvonen}; OthelloGPT \citet{hazineh2023othello}; ChessBench \citet{chessbench} &
PGN vs.\ FEN &
Causal (PGN); bidirectional (FEN) &
Transformers learn latent spatial world models; FEN often stronger in practice. \\

Scaling to GM-range &
Ruoss et al.\ \citet{ruoss2024grandmaster}; Schultz et al.\ \citet{schultz2025boardgames}; Hamara et al.\ \citet{hamara2025planning}; Ye et al.\ \citet{ye2025diffusearch} &
PGN/FEN + search or alt.\ objectives &
Mixed &
Large data/models (and search) reach superhuman strength across board games. \\

Interpretability / world models &
Toshniwal \citet{tosh}; Ivanitskiy \citet{ivanitskiy2023maze}; Spies \citet{spies2024maze}; Shai \citet{shai2024belief} &
Sequential $\rightarrow$ latent spatial &
Mostly causal &
Structured representations (legal-state tracking, causal models, belief-state geometry) emerge. \\

Masking strategies (spatial inputs) &
Amrani (block-causal images) \citet{amrani2025xtra}; Pei (relaxed V+L masks) \citet{pei2025rethinking} &
Images / V+L &
Block/relaxed causal &
Causality can be adapted rather than abandoned for spatial modalities. \\
\bottomrule
\end{tabularx}
\end{table}

\subsection{Notes and contrasts to our study}
\begin{itemize}
    \item Prior FEN work typically \emph{avoids} causal masking or uses bidirectional attention; we instead \emph{accept} causal masking on spatial inputs and find it preferable to PGN sequentialization.
    \item Our comparison holds model/data size constant across settings where possible, isolating the masking/representation choice.
    \item Results align with interpretability findings that spatial structure emerges even from sequential inputs, but we show leveraging \emph{explicit} spatial inputs with causal masking is stronger.
\end{itemize}

\subsection{Objective Function}
\label{appendix:objective}
FEN notation represents the 2D spatial orientation of the pieces in a given board state and is therefore non-causal. Because of this, each board state \( B \) is instead paired with a ground-truth label \( m^\ast \), representing the best next move according our oracle \(\Upsilon\). The objective now is to learn a model that maximizes the log probability of correctly predicting the best move \(m^\ast\) given board state \(B\):
\[\hat{m} = \arg\max_{m \in \mathcal{M}} \left[ \log P(m==m^\ast \mid B)\right]\]
With loss function:
\[
\mathcal{L} = -\sum_{j=1}^M \mathbb{I}(y_{j}==m^\ast)\log(y_{j})
\]
This loss function is the same as above, except now it is \emph{not} computed across all timesteps comprising a game, because FEN has no temporal dimension; only the target token(s) comprising \(\hat{m}\) contribute to the loss.

\section{Empirical Demonstration of Functional Complexity Difference between FEN and PGN}
\label{appendix:functionalcomplexity}
As a preliminary step ahead of our full training loop, we decided to test our theory that a mode \(\Pi_\theta^F\) trained on FEN will exhibit higher performance than an equivalent model \(\Pi_\theta^P\) trained on PGN due to the challenge of learning an accurate spatial representation from PGN data. To this end, we trained Karvonen's 50M parameter decoder-only NanoGPT on equivalently sized datasets of PGN and FEN data. We trained both models for 200,000 training steps with equal batch sizes and hyperparameter values. We removed the causal attention mask from the nanoGPT model to be trained on FEN, allowing it to attend to all tokens simultaneously, because FEN data is spatial. We then evaluated the mean cross-entropy per sample for FEN and PGN test data:\\

\[
\text{CE}_{\text{PGN}} = -\frac{1}{N}\sum_i\sum_{t \in |p_i|}\log P_{M^P_\theta}(\hat{m}_{it} | p_i) 
\]

\[
\text{CE}_{\text{FEN}} = -\frac{1}{N}\sum_i\log P_{M^F_\theta}(\hat{m}_i | f_i) 
\]

Where \(\hat{m}_i\) gives the model's predicted move for the \(i^{th}\) sample. The equation for PGN contains a second sum because a single move will contain \(t\) tokens; in FEN data, the tokenization scheme represents each move as a single token. 

\begin{figure}[H]
    \centering
    \includegraphics[width=\textwidth]{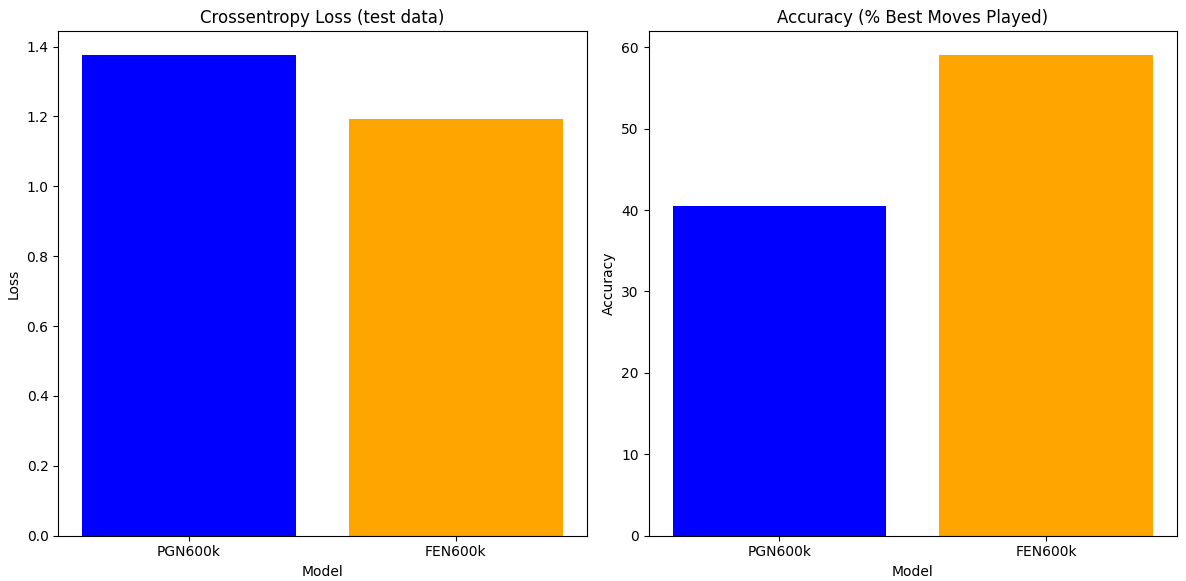}
    \caption{The total crossentropy loss per prediction (left) and accuracy in \% of top engine moves played (right) for models trained on FEN (blue) and PGN (yellow).}
    \label{fig:pgn_notation_appendix}
\end{figure}

We observed that the model trained on PGN did indeed have substantially higher loss values and lower best-move classification when compared to the model trained on FEN.

\FloatBarrier

\section{Quantifying Exposure Bias in Trained Models}
\label{appendix:exposurebias}

\subsection{Formal Definition}
Exposure bias refers to the mismatch between training, where models condition on ground-truth tokens, and inference, where they condition on their own generated tokens. Formally:

\[
  p_{\theta}\bigl(y_{t} \,\mid\, y_{<t}^{\text{model}}\bigr)
  \;\neq\;
  p_{\theta}\bigl(y_{t} \,\mid\, y_{<t}^{\text{ground-truth}}\bigr),
\]

so the model encounters different distributions at train and test time.

\FloatBarrier

\section{Training Configurations and Hyperparameters}
\label{appendix:training}

\subsection{Compute Setup}
We trained all models on two Nvidia A100 GPUs with 80GB VRAM each. Training ran for 200,000 steps using a cosine decaying learning rate, 2000 warmup steps, and gradient clipping at $\pm 1.0$. Mixed-precision training was enabled to improve memory efficiency, and gradient accumulation was used to simulate larger batch sizes. 

\subsection{Loss Curves}
\begin{figure}[H]
  \centering
  \includegraphics[width=\linewidth]{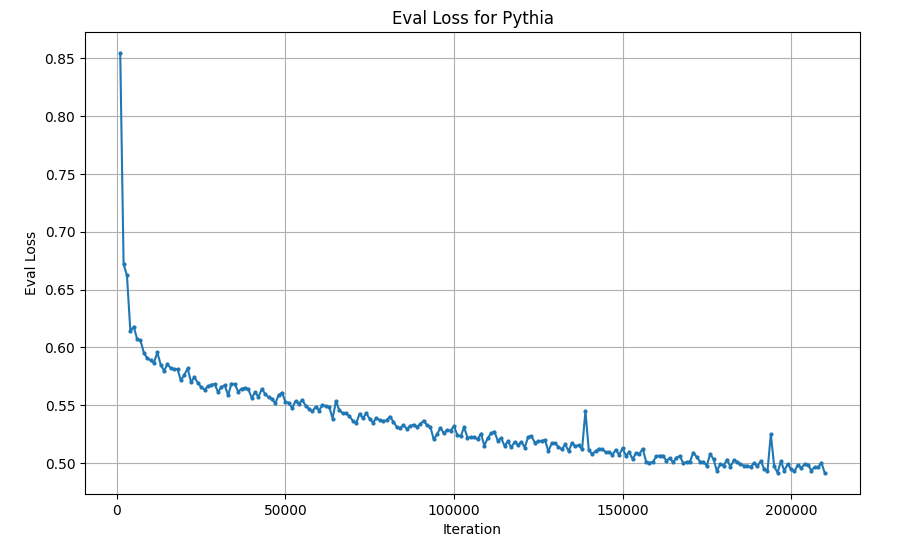}
  \caption{Pythia training loss. Greater variance due to smaller batch size.}
  \label{fig:train_pythia}
\end{figure}

\begin{figure}[H]
  \centering
  \includegraphics[width=\linewidth]{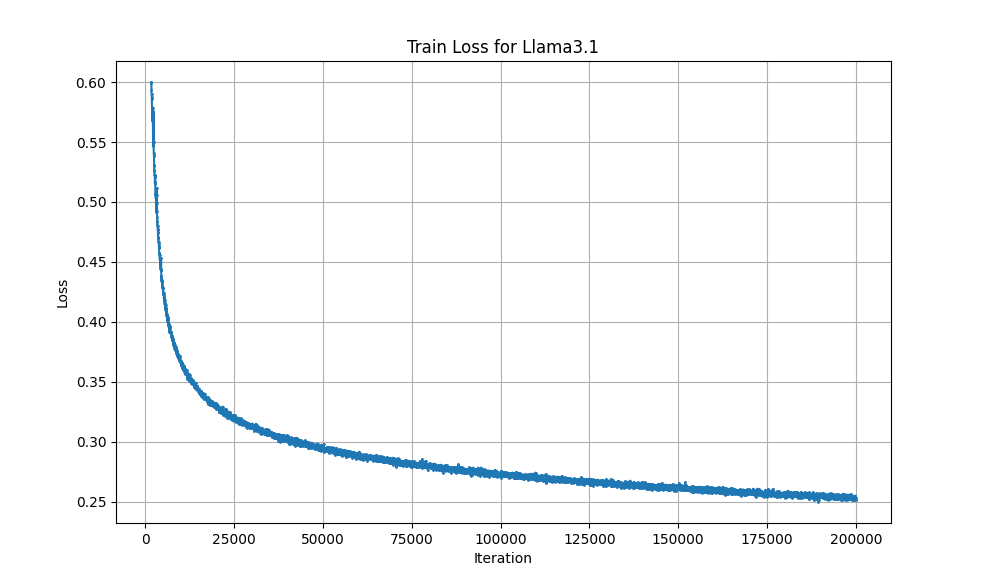}
  \caption{Llama training loss.}
  \label{fig:train_llama}
\end{figure}

Figures~\ref{fig:train_pythia}–\ref{fig:train_llama} show training loss curves.  

\subsection{Hyperparameters}
\begin{table}[t]
\centering

\caption{training hyperparameters for Llama and Pythia models.}
\vspace{1\baselineskip} 
\label{tab:hyperparameters}
\begin{tabular}{lcc}
\toprule
\textbf{Hyperparameter} & \textbf{Llama} & \textbf{Pythia} \\
\midrule
Learning rate                 & 8e-05  & 3e-04 \\
Batch size                    & 128    & 256   \\
Gradient accumulation steps   & 32     & 4\textsuperscript{*} \\
Token batch size              & 32{,}768 & 8{,}192 \\
Max iterations                & 200{,}000 & 200{,}000 \\
$\beta_1$                     & 0.9    & 0.9   \\
$\beta_2$                     & 0.95   & 0.95  \\
Warmup iterations             & 2{,}000  & 2{,}000 \\
LR decay iterations           & 200{,}000 & 200{,}000 \\
Minimum LR                    & 5e-06  & 6e-05 \\
\bottomrule
\end{tabular}
\end{table}

\subsection{Training Dynamics}
We observed that increasing effective batch size sped convergence sublinearly while increasing runtime linearly. This required balancing accuracy against compute budget. Despite this, all models converged to strong performance.  

\subsection{Experimental Setup}
We prompted our fully trained Llama and Pythia models with 12,800 board positions sampled from our test dataset. Both models generated predictions under two conditions:
1. **Teacher-forced next-token generation** (conditioning on ground-truth tokens).  
2. **Autoregressive generation** (conditioning on the model’s own predictions).  

We then compared performance on three metrics:  
- (\%) syntactically valid chess moves,  
- (\%) legal moves in the given board state,  
- (\%) best move according to Stockfish.

\subsection{Results}
We found that Llama and Pythia incurred modest penalties with respect to best-move prediction at inference time (Llama: $\sim 3\%$ drop; Pythia: $<1\%$ drop). Interestingly, both models were more likely to generate legal and valid SAN moves in autoregressive mode, suggesting they internalized structural constraints of chess notation.  

\begin{figure}[!t]
  \centering
  \includegraphics[width=\linewidth]{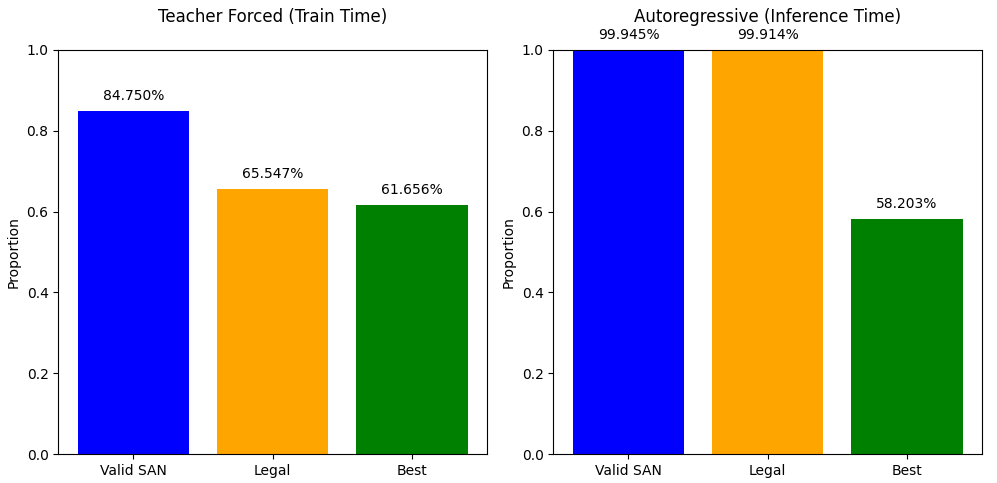}
  \caption{Llama exposure bias analysis.}
  \label{fig:llama_exposure_bias}
\end{figure}

\begin{figure}[!t]
  \centering
  \includegraphics[width=\linewidth]{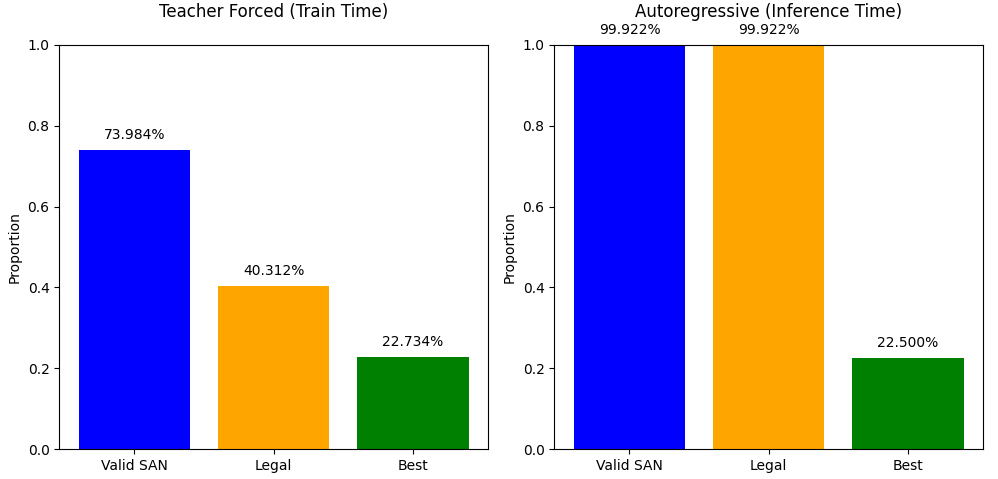}
  \caption{Pythia exposure bias analysis.}
  \label{fig:pythia_exposure_bias}
\end{figure}

\FloatBarrier

\section{Background: PGN and FEN}
\label{appendix:notation}

\paragraph{PGN (Portable Game Notation).}
A text format that records a game as a \emph{sequential} list of moves (typically in SAN), optionally preceded by metadata (event, site, players, etc.). Applying the legal move sequence from any PGN prefix deterministically produces a unique board position.

\paragraph{FEN (Forsyth–Edwards Notation).}
A \emph{spatial} snapshot of a single position using six fields:
(1) piece placement (ranks 8\(\rightarrow\)1 separated by “/”; digits count consecutive empty squares),
(2) active color (\texttt{w} or \texttt{b}),
(3) castling availability (\texttt{KQkq} subset or “-”),
(4) en passant target square or “-”,
(5) halfmove clock (for the 50-move rule),
(6) fullmove number.

\paragraph{Example.}
\texttt{rnbqkbnr/pppppppp/8/8/8/8/PPPPPPPP/RNBQKBNR w KQkq - 0 1}
denotes the initial position: White to move; all castling rights available; no en passant target; zero halfmoves; fullmove 1.

\paragraph{Relation.}
There is a deterministic mapping \(\mathcal{G}: \mathcal{P}\!\to\!\mathcal{F}\) from a legal PGN prefix \((m_0,\ldots,m_{t-1})\) to its FEN position \(f_t\), i.e., \(\mathcal{G}(p_t)=f_t\).
See Figure ~\ref{fig:pgn_notation} in the main text for a side-by-side illustration.

\section{Detailed Tokenization and Prompting Strategy}
\label{appendix:tokenization}
For FEN data, our bidirectional model used the tokenization approach from Ruoss et al., while our causal Llama model used a modified version designed to better align with pretrained tokenization. In particular, we enforced character-level tokenization to avoid ambiguous merges (e.g., ``pk'' for pawn–king) and flattened run-length encodings. We also embedded FEN strings alongside the list of legal SAN moves and the best Stockfish move in a templated prompt. This preprocessing stabilized training and improved convergence.

\textbf{Flattening FEN strings.} We flattened run-length encodings in FEN to ensure consistency (e.g., \texttt{r1bqk2r/} $\rightarrow$ \texttt{r.bqk..r/}). Periods signify empty squares.

\textbf{Character-level tokenization.} We forced the tokenizer to tokenize strictly at the character level. For example, Llama’s tokenizer would otherwise tokenize ``pk'' as a single token \texttt{[21486]}, but our approach tokenized it as \texttt{['p', 'k']} $\rightarrow$ [79, 74].

\textbf{Prompt templates.} Our final prompts embedded FEN strings, the list of legal SAN moves, and the Stockfish best move, in a templated instruction. For example:

\begin{quote}
\texttt{<|begin\_of\_text|> You are a chess grandmaster. This is the board in FEN notation: \{FEN\}. 
The legal moves are: \{List of Legal Moves\}. Which of these is the best move? 
Best move: \{Best Move\} <|end\_of\_text|>}
\end{quote}

We evaluated more than a dozen prompt variants (Appendix~\ref{appendix:prompting}), selecting the above based on zero-shot performance. We also padded sequences to handle up to 60 legal moves and promotions requiring 5 tokens, covering 99.95\% of positions in the dataset.

\FloatBarrier

\section{Loss Curves for Training}
\label{appendix:losscurves}
Note that the greater variance in the loss curve for Pythia is due to a combination of smaller batch size and the fact that we switched our logging system to log the running mean loss across training steps before starting Llama. 

\FloatBarrier

\section{Stockfish Ratings}
\label{appendix:stockfish_ratings}
\begin{table}[H]
\centering
\caption{Stockfish levels and their approximate Elo ratings used for calibration.}
\vspace{1\baselineskip} 
\label{tab:stockfish_ratings}
\begin{tabular}{cc}
\toprule
\textbf{Stockfish level} & \textbf{Elo rating} \\
\midrule
0  & 1320 \\
1  & 1467 \\
2  & 1608 \\
3  & 1742 \\
4  & 1922 \\
5  & 2203 \\
6  & 2363 \\
7  & 2499 \\
8  & 2596 \\
9  & 2702 \\
10 & 2788 \\
\bottomrule
\end{tabular}
\end{table}

\FloatBarrier

\section{Elo Rating Evaluation}
\label{appendix:elo}

\subsection{Overview of the Elo System}
Chess players' skill is typically quantified using the Elo rating system. Elo ratings are not absolute, but relative to those of other players, and are determined by comparing expected vs. actual performance.  
As of 2025, the world's strongest human player is Magnus Carlsen, with an Elo rating of 2831 \citet{FIDE}. Stockfish, the world's strongest chess engine, has an estimated Elo rating of 3642 \citet{ccrl}.  

\subsection{Evaluation Protocol}
To estimate Elo ratings, we played each model in a large-scale simulated tournament against attenuated Stockfish agents with known approximate strengths (Levels 0–10, see Table~\ref{tab:stockfish_ratings}). Each model played:
- 1000 games per Stockfish level (500 as White, 500 as Black).  
- The PGN-trained model played only as White, due to dataset limitations.  
- To ensure diverse game states, we used an opening book at the start of each game.  
- Moves were generated with temperature-based stochastic sampling. If more than 5 consecutive illegal moves were produced, the game was terminated and counted as a loss. (In practice, none of our models forfeited games this way.)

\subsection{Elo Computation}
Elo ratings were computed relative to the known ratings of the Stockfish agents. The Elo update formula was:

\[
R_{\text{new}} = R_{\text{current}} + K \cdot (W - N \cdot E)
\]

where \(W\) is the number of wins, \(N\) the total games, \(E\) the expected win rate, and $K=16$. The expected win rate was defined as:

\[
E = \frac{1}{1 + 10^{\frac{R_{\text{opponent}} - R_{\text{current}}}{400}}}.
\]

\subsection{Results}
Our Llama model shows grandmaster-level performance, winning or drawing more than half of its games against Stockfish 7 (approx. Elo 2500), and even managing winning results in 30\% of its games against Stockfish 10. Its estimated Elo rating based on Stockfish evaluations is 2630, stronger than approximately 90\% of human grandmasters, only marginally weaker than the estimated Elo rating of 2680 associated with our bidirectional model, and substantially stronger than our smaller causal LLM, which had an estimated Elo of 2000, as well as GPT 3.5-Turbo, which was estimated to have an Elo rating around 1750 \citet{acher}. 

\begin{figure}[H]
    \centering
    \includegraphics[width=\linewidth]{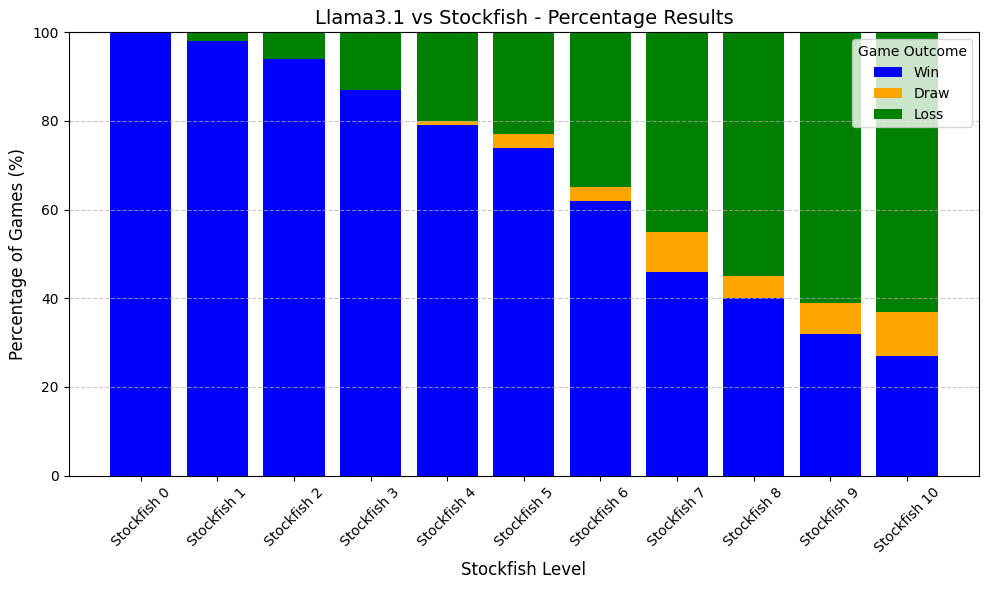}
    \caption{Llama wins out of 1000 against Stockfish 0–10.}
    \label{fig:llama_vs_stockfish}
\end{figure}

\begin{figure}[H]
    \centering
    \includegraphics[width=\linewidth]{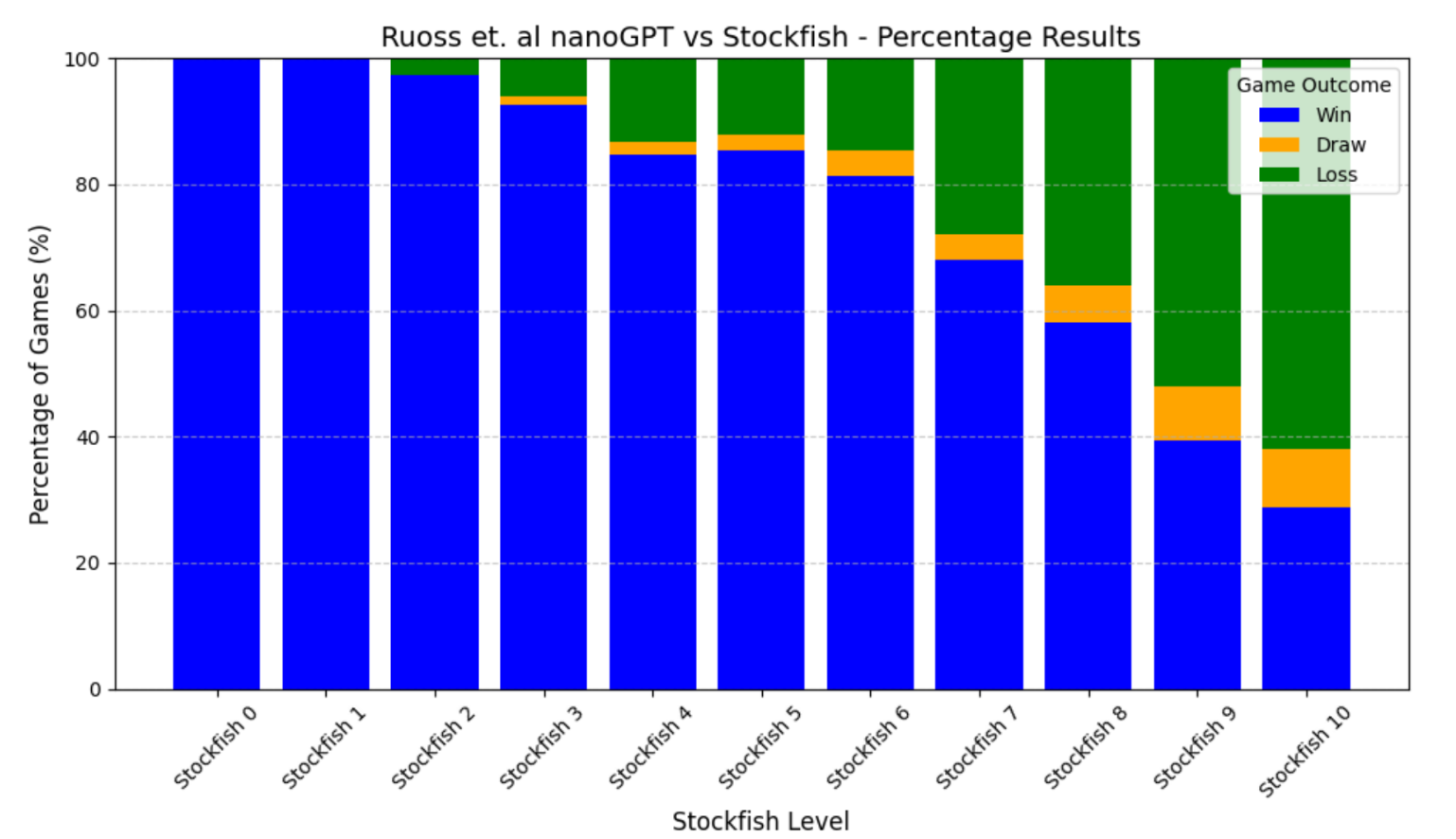}
    \caption{Ruoss et al. wins out of 1000 against Stockfish 0–10.}
    \label{fig:ruoss_vs_stockfish}
\end{figure}

\begin{figure}[H]
    \centering
    \includegraphics[width=\linewidth]{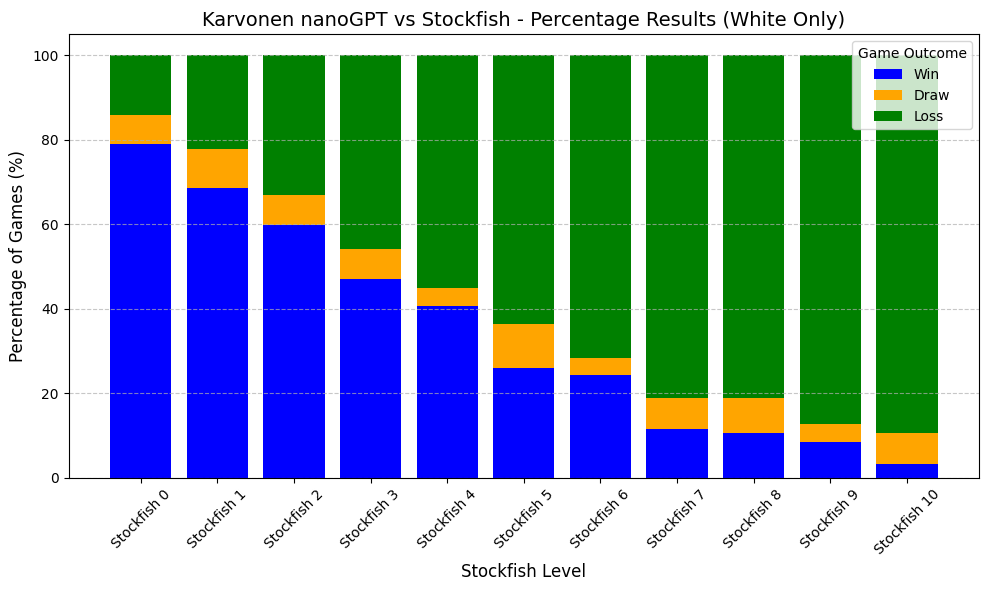}
    \caption{Karvonen wins out of 1000 against Stockfish 0–10 (White games only).}
    \label{fig:karvonen_vs_stockfish}
\end{figure}

Our Pythia model performs relatively poorly, winning games only against the first six Stockfish agents and losing heavily overall.  

Noever et al.’s GPT-2 model failed to complete any games against Stockfish 0 (Elo 1300), losing every game due to repeated illegal moves. We ascribe this to short training (33,000 steps for 774M parameters), lack of tokenizer modifications, and instability in its published training loss curve.  

Karvonen's nanoGPT showed strong performance as White but lost all games as Black. This asymmetry stems from dataset bias, since training games involved a strong engine always playing as White. Its estimated Elo is 2000 (White-only) or 1301 (overall), consistent with Karvonen’s reported results.  

Ruoss et al.’s model achieved an Elo rating of 2682, marginally higher than our Llama. It substantially outperformed against Stockfish 5–7, but converged with our model’s performance at Stockfish 9–10. Their published result was 2895, but we only trained for 200K steps vs. their 1M, explaining the gap.

\subsection{Interpretation}
While these Elo ratings provide a relative benchmark against calibrated Stockfish levels, they are not directly comparable to official FIDE Elo ratings for human players. Nonetheless, they place our model’s performance in a range typically associated with grandmaster-level play.

\section{Distributional analysis of number of legal moves in board position}
\label{appendix:nummoves}
\begin{figure}[H]
    \centering
    \includegraphics[width=\textwidth]{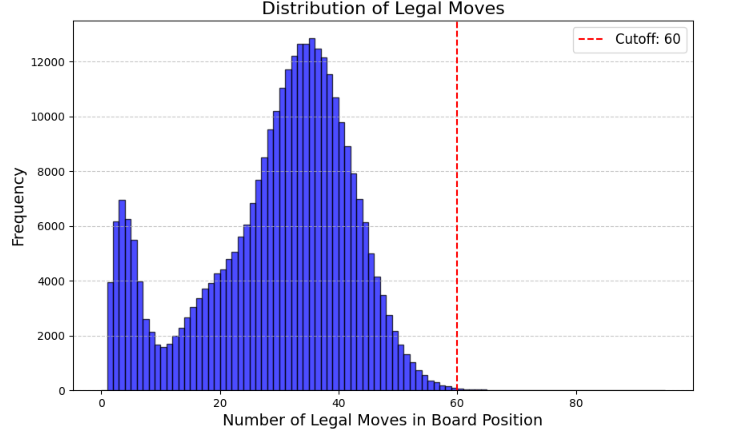}
    \caption{Distribution of number of legal moves in a sample of our dataset. The secondary peak along the left tail of the distribution is due to positions in which the king is in check and other pieces may not move.}
    \label{fig:nummoves}
\end{figure}

\FloatBarrier

\section{Zero Shot Prompting Ablation Study}
\label{appendix:prompting}

\subsection{Prompting Ablations}
Prompting strategy significantly affected zero-shot performance. Prompts framing the model as a “chess grandmaster” and constraining outputs to the provided legal moves yielded the best results. Full ablation results, prompt templates, and comparative figures are included in Appendix~\ref{appendix:prompting}.

\begin{figure}[!t]
    \centering
    \includegraphics[width=\textwidth]{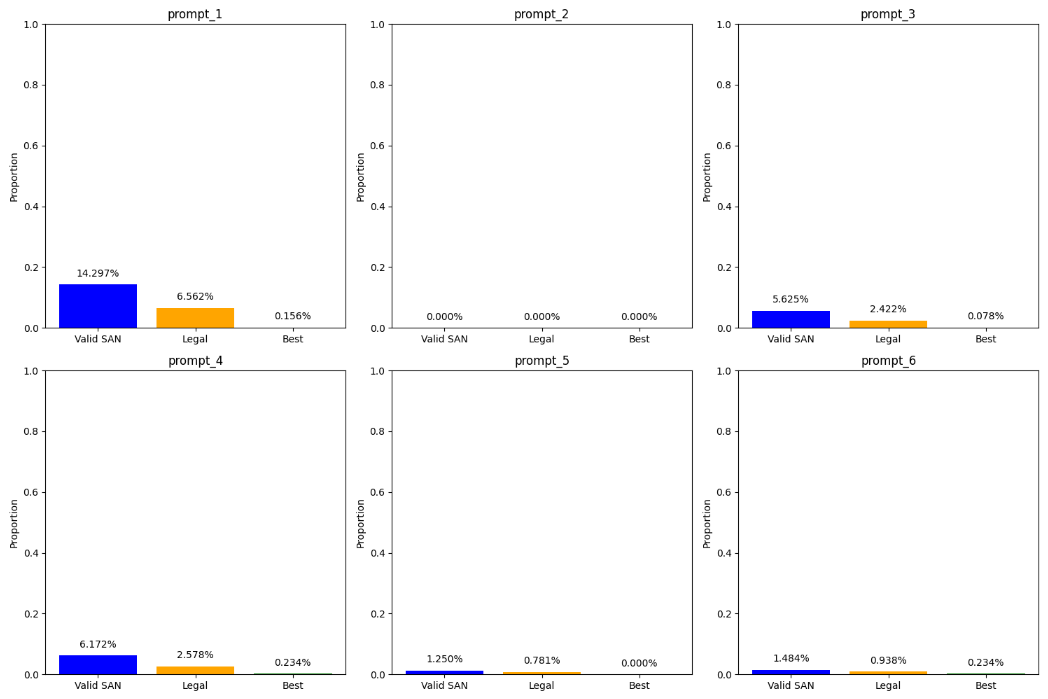}
    \caption{Baseline performance for the six best prompts (Pythia). 
    Valid SAN denotes exemplars where Pythia output a move that could 
    have been legal on any board state; Legal denotes exemplars where 
    the move was legal in the current board state; Best denotes exemplars 
    where the move matched the engine's best move.}
    \label{fig:pythia_prompt_perf}
\end{figure}

\begin{figure}[!t]
    \centering
    \includegraphics[width=\textwidth]{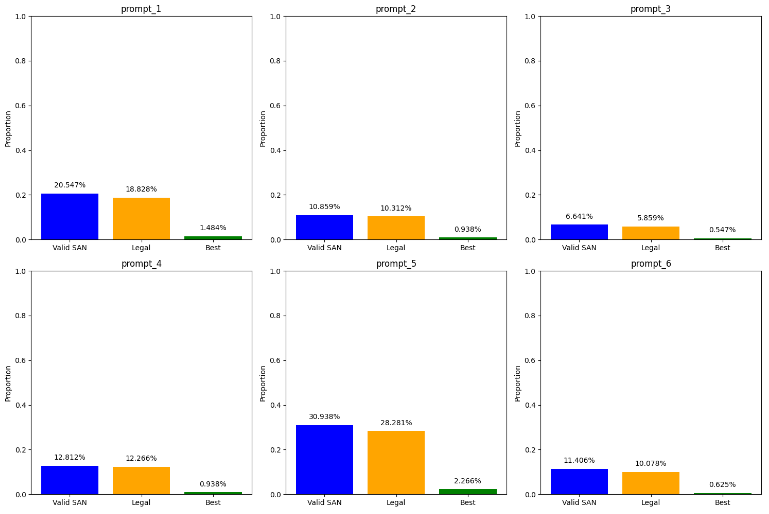}
    \caption{Baseline performance for the six best prompts (Llama).}
    \label{fig:llama_prompt_perf}
\end{figure}

Top 6 Performing Prompts (ordered according to name in charts above, not according to performance): 

\[
\text{\parbox{\linewidth}{Prompt 1 = \texttt{"<|begin\_of\_text|> You are a chess grandmaster. This is the board position in FEN notation: \{FEN tokens\}. The legal moves are: \{List of Legal Moves in SAN\}. Which of these is the best move? Best move: \{Best Move in SAN\} <|end\_of\_text|> <|pad|> <|pad|> <|pad|> <|pad|> …"}}}
\]
\[
\text{\parbox{\linewidth}{Prompt 2 =  \texttt{"<|begin\_of\_text|> [White 'Magnus Carlsen'] [Black 'Stockfish'] Board position: \{FEN tokens\}, Legal Moves: \{List of Legal Moves in SAN\}, Best Move: \{Best Move\} <|end\_of\_text|> <|pad|> <|pad|> <|pad|> <|pad|> …"}}}
\]
\[
\text{\parbox{\linewidth}{Prompt 3 = \texttt{"<|begin\_of\_text|> You are a chess grandmaster. This is the board in fen (Forsyth-Edwards notation). It is your move: \{FEN tokens\}. Please select the best move from this list: \{List of Legal Moves in SAN\}.. Please ONLY PLAY MOVES LISTED HERE. ANY move not in here is illegal. Best move:  \{Best Move in SAN\} <|end\_of\_text|> <|pad|> <|pad|> <|pad|> <|pad|> …"}}}
\]
\[
\text{\parbox{\linewidth}{Prompt 4  = \texttt{"<|begin\_of\_text|> You are analyzing a competitive chess game. The current board position is represented in FEN notation: \{FEN tokens\}. The legal moves available are: \{List of Legal Moves in SAN\}.. Based on the position, decide which move is the best. Best move: \{Best Move in SAN\} <|end\_of\_text|> <|pad|> <|pad|> <|pad|> <|pad|> …"}}}
\]
\[
\text{\parbox{\linewidth}{Prompt 5 = \texttt{"<|begin\_of\_text|> [FEN '\{FEN tokens\}'] Legal Moves: \{List of Legal Moves in SAN\}. Based on the current board, determine the best move from the provided options. Best Move: \{Best Move in SAN\} <|end\_of\_text|> <|pad|> <|pad|> <|pad|> <|pad|> …"}}}
\]

\[
\text{\parbox{\linewidth}{Prompt 6 =  \texttt{"<|begin\_of\_text|> As a world-class chess engine, your task is to analyze the following board position and select the best move. Board in FEN: \{FEN tokens\}. Legal moves available: \{List of Legal Moves in SAN\}}}}
\]

\FloatBarrier

\end{document}